\documentclass[9pt,sigconf,nonacm]{acmart}

\usepackage{multirow}
\usepackage{bm}
\usepackage{textcomp}
\usepackage{xcolor}
\usepackage{color}
\usepackage{color,soul}
\usepackage{underscore}
\usepackage{pifont}
\usepackage{caption}
\usepackage{bm}
\usepackage{url}
\usepackage{algorithmic}
\usepackage[ruled,vlined]{algorithm2e}
\usepackage{graphicx}
\usepackage{textcomp}
\usepackage{xcolor}
\usepackage{multirow}
\usepackage{amsmath}
\usepackage{geometry}
\usepackage{tikz}
\usepackage{enumitem}
\usepackage{fancyhdr}
\usepackage{hyperref}

\AtBeginDocument{%
  \providecommand\BibTeX{{%
    \normalfont B\kern-0.5em{\scshape i\kern-0.25em b}\kern-0.8em\TeX}}}

\newcommand*\circled[1]{\raisebox{.4pt}
                    {\tikz[baseline=(char.base)]{
            \node[shape=circle,draw,inner sep=1pt, style={fill=black, text=white}, scale=0.75] (char) {\textbf{#1}};}}}
            
\setcopyright{none}
\begin{document}



\title{H2H: Heterogeneous Model to Heterogeneous System Mapping with Computation and Communication Awareness}








\author{ \normalsize Xinyi Zhang$^1$, Cong Hao$^2$, Peipei Zhou$^1$, Alex Jones$^1$, and Jingtong Hu$^1$ \\
$^1$University of Pittsburgh, $^2$Georgia Institute of Technology \\
\{xiz173, peipei.zhou, akjones, jthu\}@pitt.edu, callie.hao@ece.gatech.edu \\}

\renewcommand{\shortauthors}{}

\begin{abstract}
The complex nature of real-world problems calls for heterogeneity in both machine learning (ML) models and hardware systems. 
The heterogeneity in ML models comes from multi-sensor perceiving and multi-task learning, i.e., multi-modality multi-task (MMMT), resulting in diverse deep neural network (DNN) layers and computation patterns. 
The heterogeneity in systems comes from diverse processing components, as it becomes the prevailing method to integrate multiple dedicated accelerators into one system.
Therefore, a new problem emerges: \textit{heterogeneous model to heterogeneous system mapping} (\textbf{H2H}).
While previous mapping algorithms mostly focus on efficient computations, in this work, we argue that it is indispensable to consider computation and communication simultaneously for better system efficiency. We propose a novel H2H mapping algorithm with both computation and communication awareness;
by slightly trading computation for communication, the system overall latency and energy consumption can be largely reduced.
The superior performance of our work is evaluated based on MAESTRO modeling, demonstrating 15\%-74\% latency reduction and 23\%-64\% energy reduction  compared with existing computation-prioritized mapping algorithms.
Code is publicly available at {\textcolor{cyan}{ \url{https://github.com/xyzxinyizhang/H2H}}}.

\end{abstract}

\maketitle
\vspace{-0.2cm}
\section{Introduction}\label{sec:intro}



\begin{figure}[t]
  \centering
  \includegraphics[width=0.46\textwidth]{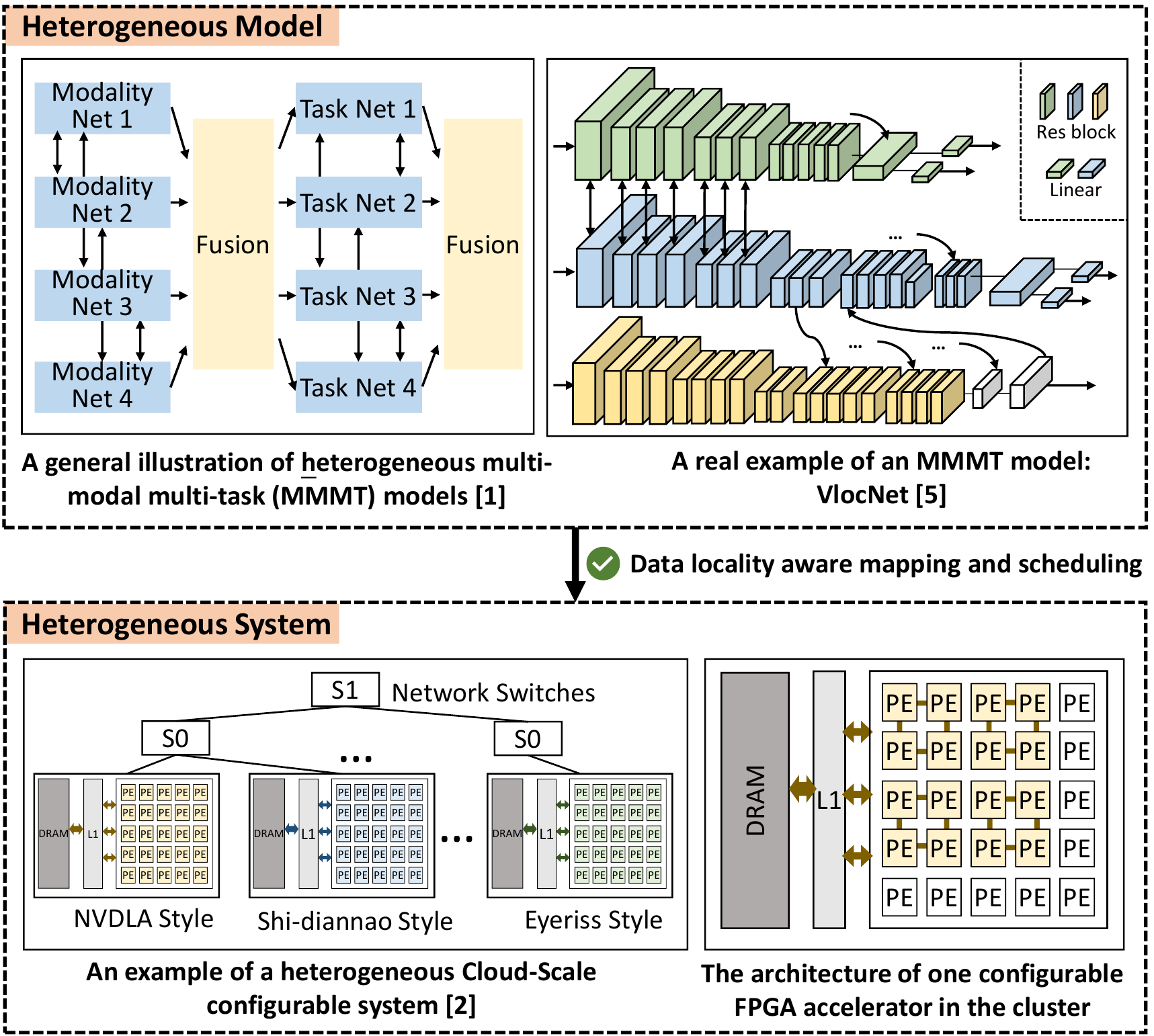}
  \caption{The overview of H2H mapping and scheduling: from heterogeneous models to heterogeneous system.}\label{Fig:H2H}
\end{figure}

As DNNs are applied in more and more complicated tasks, both machine learning (ML) models and hardware acceleration systems call for \textbf{heterogeneity} \cite{hao2021software,fowers2018configurable} to address emerging challenges. 
\underline{First}, ML algorithms are evolving from handling single-modality single-task to multi-modality multi-task (MMMT)~\cite{hao2021software}.
For instance, in the recommendation system, visual and textual data are jointly learned in a multi-modality fashion~\cite{iqbal2018multimodal}; 
in AR/VR applications, image, gesture, and speech are jointly learned~\cite{mehler2018vannotator}. 
Such changes result in increasingly complicated ML models with larger size and complex inter-block connections.
The top of Fig.~\ref{Fig:H2H} depicts 
heterogeneous ML models as well as a real-life model, VlocNet~\cite{vlocnet}, for semantic visual localization.
\underline{Second}, recent advanced systems are introducing great heterogeneity by integrating diverse acceleration components with different capabilities to pursue low latency and high energy efficiency. For example, at the System-on-Chip (SoC) level, Xilinx Versal \cite{gaide2019xilinx}, Nvidia Xavier \cite{ditty2018nvidia}, and Tesla FSD \cite{talpes2020compute} integrate various processing components on a single chip. At the cloud level, Microsoft's Brainwave \cite{fowers2018configurable} and AWS \cite{awsspeed} are composed of multiple FPGAs which can be flexibly reconfigured.
In this paper, we target a cloud-scale multi-FPGA~\cite{fowers2018configurable} as the target system architecture, shown in the bottom part of Fig.~\ref{Fig:H2H}, where each leaf device is an FPGA that can be configured as an arbitrary accelerator.

\begin{figure}[htp]
  \centering
  \includegraphics[width=0.47\textwidth]{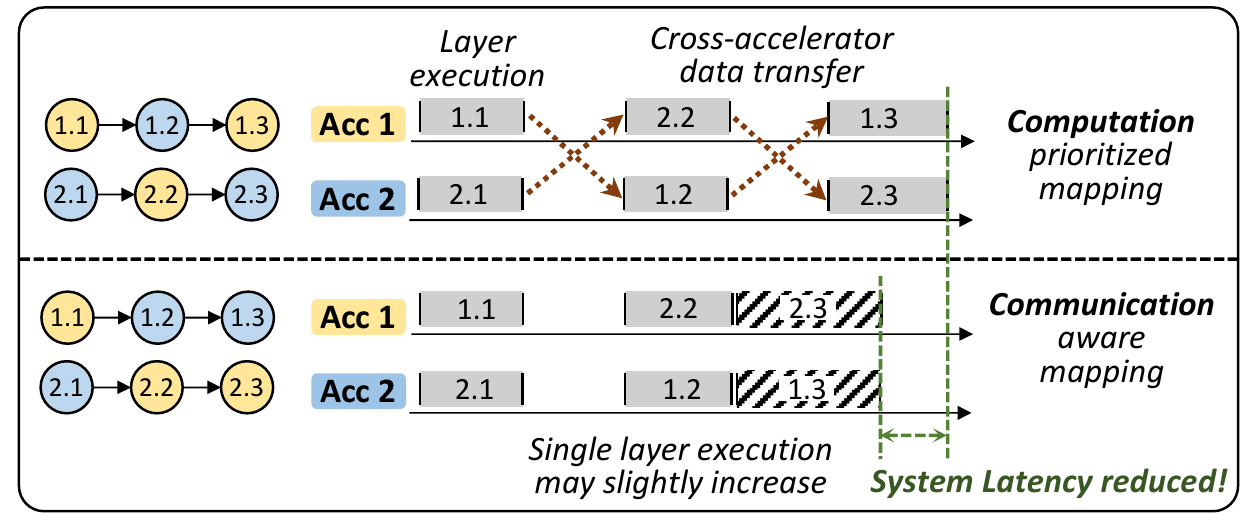}
  \caption{An example of communication-prioritized  mapping and communication-aware mapping.}\label{Fig:toyexample}
\end{figure}

We refer to 
mapping and scheduling a heterogeneous ML model onto a heterogeneous system as \textbf{H2H}.
The H2H problem is non-trivial because of the following complexities.
(1) \textbf{Computation Awareness}. The heterogeneous MMMT model components can vary largely in terms of layer type, layer shape, and data dimension. For instance, an MMMT model can be composed of convolution (Conv), fully connected (FC), long short-term memory (LSTM), and transformer layers, with significantly different dataflow patterns and preferred accelerator architectures~\cite{kwon2021heterogeneous}. A first challenge is revealed, how to map layers to the desirable accelerators, suitable for their computation patterns.
(2) \textbf{Communication Awareness}. Computation-prioritized mapping does not necessarily lead to global best performance if communication, i.e., data transfer across different accelerators, is ignored. 
Fig.~\ref{Fig:toyexample} demonstrates the difference between computation-prioritized mapping and communication-aware mapping: the former maps each layer purely based on its preferable dataflow pattern, while the latter slightly sacrifices computation efficiency but in turn reduces the overall system latency by avoiding expensive data transfer.
Thus, a second challenge emerges, as modern MMMT models have more complex dependencies (e.g., the cross-layer connections in VlcoNet~\cite{vlocnet} in Fig.~\ref{Fig:H2H}), the data transfer overhead and optimization difficulty become exaggerated.

Existing mapping algorithms for DNNs are primarily prioritized based on computation. 
For instance, Fowers et al. \cite{fowers2018configurable} improve a single accelerator computation efficiency by augmenting the dataflow, but do not discuss system-level cross-accelerator communication.
Chen et al. \cite{chen2019cloud} map DNN layers to different accelerators to fully utilize DSP and RAM resources;
Kwon et al. \cite{kwon2021heterogeneous} propose the state-of-the-art mapper, which improves the convolution efficiency by 65\% by mapping Conv layers to different styles of accelerators such as Eyeriss, NVDLA, and Shi-diannao \cite{chen2016eyeriss,NVIDIA,du2015shidiannao}. The data transfer overhead, however, is not considered.




To address the discussed challenges and limitations, in this work, we propose the first \textbf{H2H mapping algorithm with both computation and communication awareness}, aiming at system-level formulation, modeling, and optimization.
We first formulate the H2H problem using two graphs to depict the model layer dependency and accelerator execution order.
We then build a system-level modeling infrastructure to model arbitrary heterogeneous systems, based on each accelerator's computation and communication performance models.
Guided by the infrastructure, we propose an H2H mapping algorithm, aiming to largely reduce system overall latency and energy.
We summarize our contributions as follows:

\begin{itemize}[leftmargin=10pt]

\item \textbf{H2H problem formulation and system modeling.}
We formulate the H2H problem as two directed graphs to describe layer dependency of the heterogeneous model and accelerator execution order in the heterogeneous system.
We then develop a configurable system-level infrastructure based on MAESTRO~\cite{maestro}, which takes arbitrary accelerators with user-defined performance models in a plug-in manner to obtain system latency and energy.

\item \textbf{H2H mapping algorithm}{
We propose an H2H mapping algorithm with both computation and communication awareness, including computation-prioritized mapping, weight locality and activation transfer optimization, and data locality aware remapping. An optimized mapping can be found within seconds.}

\item \textbf{Performance evaluation.}
We comprehensively evaluate our mapping algorithm on the infrastructure using 6 real-world heterogeneous DNN models on a heterogeneous system composed of 12 off-the-shelf accelerators. We achieve 15\% to 74\% latency reduction and 23\% to 64\% energy reduction compared with existing computation-prioritized mapping algorithms.

\end{itemize}

\section{BACKGROUND AND MOTIVATION}

\noindent
\textbf{Heterogeneous ML Model.}
Modern real-life ML applications usually involve multiple inputs and handle multiple tasks simultaneously, i.e., multi-modal multi-task (MMMT)~\cite{hao2021software}.
Multi-modal models aim to process and relate information from multiple sources to capture the correspondences between modalities, while multi-task models aim to efficiently enforce cross-task information exchange for improving performance and robustness.
Such MMMT models expose great heterogeneity with various layer types, complex inter-layer connections, and data exchange.
In practice, convolution layers are widely seen in vision tasks; FC layers are mostly used in small-scale data extraction; LSTM/Transformer layers are prevalent in text and language tasks \cite{guo2019dl}. Different applications can have largely varied layer types with different preferable computing patterns, accelerator designs, and data transfer cost. 


\noindent
\textbf{Heterogeneous System.}
A configurable multi-FPGA system is one representative heterogeneous system, since each FPGA can be flexibly configured with different accelerator architectures, such as Eyeriss \cite{chen2016eyeriss}, NVDLA \cite{NVIDIA}, and Shi-diannao \cite{du2015shidiannao}.
Microsoft FPGA datacenter \cite{fowers2018configurable} is a good heterogeneous example as shown in Fig.~\ref{Fig:H2H} (bottom). 
Heterogeneous systems can largely improve computation efficiency, because a certain accelerator is typically optimized only for a subset of ML layers, and different layers are expected to be mapped to their specific accelerators.



\noindent
\textbf{Motivation.}
\label{sec:motivation}
Given the complexity of ML models and heterogeneous systems, it is non-trivial to find a good H2H mapping that can effectively balance \textbf{computation} and \textbf{communication}.
\underline{First}, DNN accelerators are highly specialized for certain dataflows. For instance, NVDLA \cite{NVIDIA} optimizes convolution channel-wise parallelism, while Shi-diannao~\cite{du2015shidiannao} optimizes feature-map-wise parallelism. Model layers should be mapped to the accelerators with preferable computation patterns to reduce computation latency. Most existing approaches highly prioritize the computation pattern but usually ignore cross-layer communication~\cite{kwon2021heterogeneous}.
\underline{Second}, there are also communication-prioritized mapping algorithms \cite{taura2000heuristic} by forming task clusters and assigning a cluster to a processor. However, this may largely hurt the computing efficiency since the tasks within the same cluster do not necessarily run efficiently on the same accelerator.
Meanwhile, the heavy cross-layer dependency (\textit{cross-talk}) in the heterogeneous models may also lead to ineffective clustering.
\underline{Third}, existing mapping algorithms lack the formulation of DNN models and system-level information such as accelerators' architecture and dataflow. Without hardware awareness, existing algorithms cannot be directly and efficiently applied.
Therefore, a hardware-aware mapping algorithm that considers both computation and communication simultaneously is needed.

The cornerstone for communication reduction is \textbf{data locality} by efficiently utilizing accelerators' local DRAM.
In a multi-FPGA system, each FPGA is equipped with a local DRAM, which can be utilized to store model weights and to buffer intermediate activations of two adjacent layers to reduce cross-FPGA data movement. 
The challenge is that the computation-prioritized mapping can achieve best efficiency per FPGA, but the overall performance may be compromised due to the transmission cost (and vice versa). 
Therefore, we propose an H2H mapping algorithm, which can jointly consider the benefit of high data locality and suitable computation patterns.

\begin{footnotesize}
\begin{table}[]
\tabcolsep 5.5pt
\renewcommand\arraystretch{1.2}
\captionsetup{font=small}
\caption{System Performance Modeling Parameters}
\label{table:layerpara}
\begin{tabular}{c|c|c}
\toprule
\textbf{Acc Type} & \textbf{{Parameters}} & \textbf{Explanation} \\ \midrule
Conv       & \textless{}N, M, R, C, K, S\textgreater{}           & \begin{tabular}[c]{@{}c@{}}ofm\_channel\_num, ifm\_channel\_num, \\ ofm\_height, ofm\_width, kernel\_size, stride\end{tabular} \\ \hline
FC        & \textless{}N,M\textgreater{}                        & in\_features, out\_features                                                                                                            \\ \hline
LSTM      & \textless{}N,H,L\textgreater{}                      & in\_size, hidden\_szie, layers                                                                                                         \\ \hline

-- & $BW_{acc}$ & accelerator-to-host bandwidth \\ \hline
-- & $M_{acc}$ & local DRAM size \\\bottomrule
\end{tabular}
\end{table}
\end{footnotesize}

\section{system formulation}


\noindent
\textbf{Heterogeneous ML Model}.
As shown in Fig. \ref{Fig:H2H}, a heterogeneous model has complicated dependencies especially for cross-talk connections. It is natural to formulate such a model as a directed graph $G_{model} = (V, E)$, where the vertices represent the layers and the edges represent the dependencies.
In $G_{model}$, each node holds layer information such as Conv, FC, LSTM, etc., as well as their data dimension (e.g., feature map size).
We consider three types of popular accelerators with the layer parameters
summarized in Table \ref{table:layerpara}.




\noindent
\textbf{Heterogeneous System}.
We also formulate the multi-FPGA system as a directed graph $G_{sys} = \{G_{Acc_i}\}$, where each sub-graph $G_{Acc_i}$ is a computation graph 
representing the layers' execution scheduling on the $i$-th FPGA accelerator $Acc_i$.
Initially, each graph $G_{Acc_i}$ is empty without any mapping.
An example of $G_{model}$ and initial $G_{sys}$ with three initial empty $Acc_i$ are shown in Fig.~\ref{Fig:algsteps}'s input block.
After mapping, each $G_{Acc_i}$ will be composed of nodes from $G_{model}$ in their execution order.

In this work, we consider a multi-FPGA system proposed in~\cite{fowers2018configurable}, where each FPGA is connected to a host node via Ethernet switches, enabling FPGA-to-FPGA and FPGA-to-host communication. The host node distributes data to each FPGA's local DRAM memory, whose capacity typically ranges from 512 MB to 8 GB \cite{guo2019dl} and is usually used as additional buffers to mitigate the scarcity of FPGA on-chip memory.
The Ethernet speed ranges from 1 G to 10 G Ethernet (0.125 to 1.25 GB/s)  in cloud-FPGA~\cite{awsspeed}, and FPGA local DRAM speed ranges from 6.4 GB/s to 460 GB/s \cite{riley2019basic}. 
We consider two most important system-level parameters, accelerator-to-host bandwidth and local DRAM size, denoted by $BW_{acc}$ and $M_{acc}$, respectively, as shown in Table~\ref{table:layerpara}.

\setlength{\skip\footins}{0.1cm}
\noindent
\textbf{System Performance Model}.
We model the overall heterogeneous system performance at two levels: individual accelerator, and the overall system.
\underline{First}, for individual accelerator, there are plenty of analytical models of different designs, so we directly adopt the performance models from existing literature, denoted by $P_{acc}$.
For each accelerator, we consider the following configurable parameters:
(1) $BW_{acc}$, the accelerator to main memory bandwidth;
(2) $M_{acc}$, the local DRAM size;
(3) $Layer_{para}$, other layer parameters as shown in Table~\ref{table:layerpara}.
For instance, the analytical model for the accelerator proposed in 
\cite{zhang2015optimizing} can be expressed as
$P_{Acc} $<$M_{acc}, Layer_{para}$> with its loop tiling setting
<$R_{wei}, D_{type}, F_{acc}, BW_{dram}, Tm, Tn, Tr, Tc$>. 
\underline{Second}, for the system-level performance model, we 
modify the MAESTRO for the target multi-FPGA system by allowing customizing the accelerator-to-host bandwidth as $BW_{acc}$, which is also configurable by users.



\vspace{-0.4cm}
\section{proposed H2H mapping algorithm}
\begin{figure*}[t]
\vspace{-20pt}
 \setlength{\belowcaptionskip}{-0.2cm}
  \centering
  \includegraphics[width=0.82\textwidth]{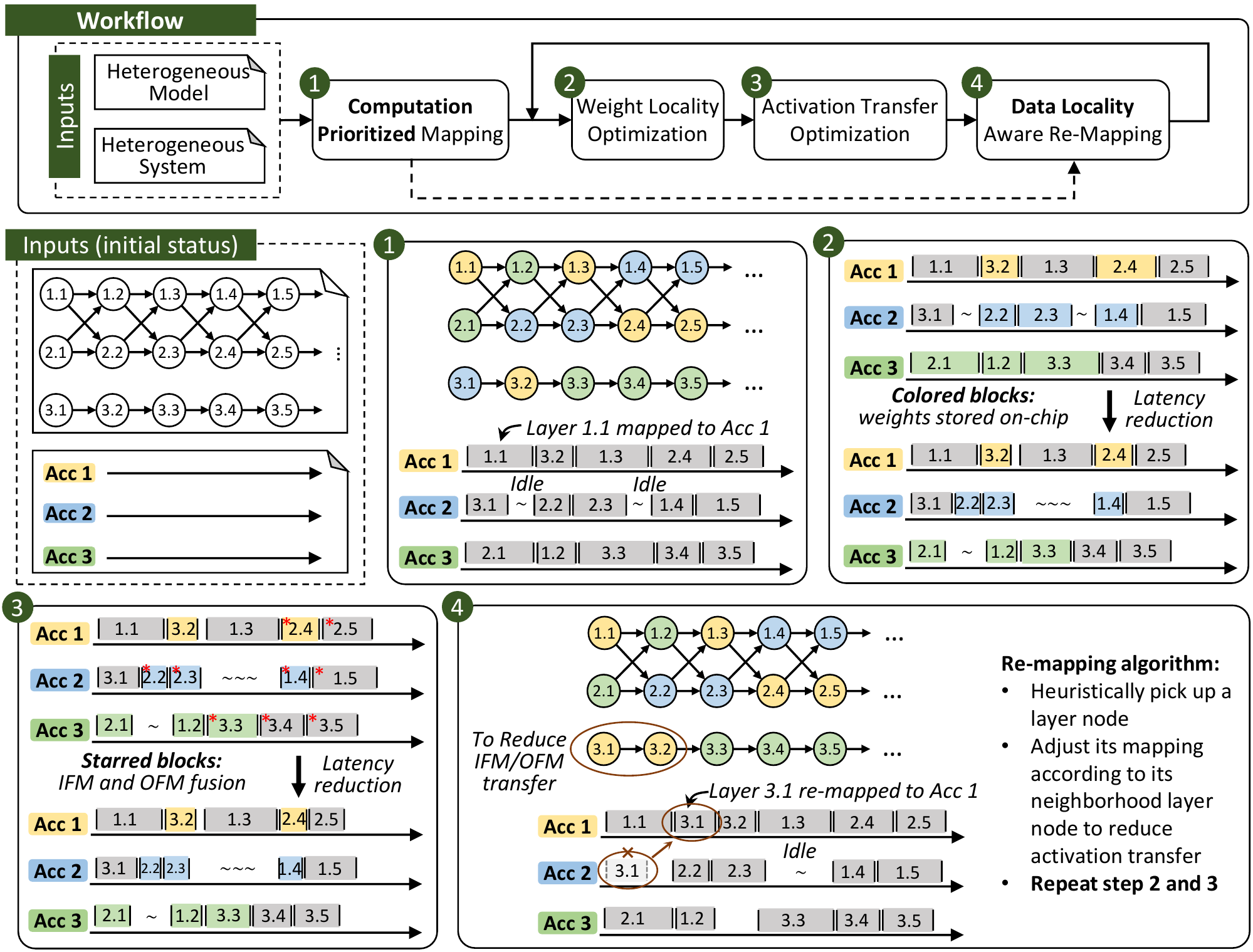}
  \caption{H2H mapping algorithm visualization. It includes 4 major steps: (1) computation-prioritized mapping; (2) weight locality optimization; (3) activation transfer optimization; (4) data locality aware remapping.}\label{Fig:algsteps}
\end{figure*}

In this section, we discuss our proposed computation and communication aware H2H mapping algorithm, according to our analytical model based infrastructure.
As shown in Fig.~\ref{Fig:algsteps} top, there are four steps.
\circled{1} \textbf{Computation-prioritized mapping}. The heterogeneous ML model is mapped at layer-granularity, that each layer is mapped to the accelerator that best fits its computation dataflow, ignoring all data movement optimizations (i.e., zero data locality). 
\circled{2} \textbf{Weight locality optimization}. Since each accelerator has its own local DRAM, we buffer part of the weights in the memory to maximally avoid weight data movement.
\circled{3} \textbf{Activation transfer optimization}. If two adjacent layers are mapped to the same accelerator, their intermediate activation, i.e., the output/input feature maps (OFM/IFM), will no longer need to transfer and thus latency can be reduced.
\circled{4} \textbf{Data locality aware re-mapping}. This step explores the trade-off between computation and communication, aiming to largely reduce communication cost with slight computation efficiency degradation, which still results in overall performance improvement. The H2H algorithm flow is shown in Algorithm \ref{alg:init}. It takes $G_{model}$ and $P_{Acc_i}$ as inputs, and produces a mapped and scheduled solution ($G_{model}^*$, $G_{sys}^{*}$) with modeled system latency and energy ($Sys_{latency}$, $Sys_{energy}$). 

\subsection{Computation-prioritized Mapping}
In the first step, we perform 
\texttt{Computation_Prioritized_Mapping}.
It assigns the model layers to the accelerators that result in the best computation performance by assuming zero local DRAM without any data locality at the accelerator.
We use performance model $P_{Acc_i}$ to estimate the latency of a layer executing on the $i$-th accelerator, and assume that all the weights and intermediate results go to the main memory of the host node.
To obtain system latency, the layer scheduling on each accelerator must be determined.
To guarantee a valid scheduling considering layer dependencies especially across multiple sub-models, the algorithm determines the mapping and scheduling iteratively.
In every iteration, it selects all the nodes without predecessors from $G_{model}$ as a group, enumerates all possible mappings within the group (multiple nodes can be mapped to one or more accelerators), and selects the one that results in the smallest system latency increment.

An example is shown in Fig.~\ref{Fig:algsteps} \circled{1}, where the color of the nodes represents which accelerator it is mapped to. The gray blocks represent the accelerator execution, where idle periods are introduced by layer dependency.
Note that, in this step, we assume zero local DRAM, so that the latency values include both computation and communication: layer computation, weight transfer from the main memory, and IFM/OFM transfer from/to the main memory.

\setlength{\textfloatsep}{0.2cm}
\begin{algorithm}[t]
\LinesNumbered 
\footnotesize
\SetKwProg{try}{try}{:}{}
\SetKwProg{accept}{accept}{:}{end}
\KwIn{$G_{model}$, $P_{Acc_i}$}
\KwOut{$G_{model}^*$, $G_{sys}^{*}$ = \{$G_{Acc_i}^{*}$\}, $Sys_{latency}$, $Sys_{energy}$}

\SetKwFunction{Fone}{Computation\_Prioritized\_Mapping}
\SetKwProg{Fn}{Function}{:}{}
\Fn{\Fone{}}{
    \For{nodes in $G_{model}$ without predecessors}{
        Enumerate all possible mappings based on $P_{Acc_i}$ \\
        Choose the mapping with minimum $\Delta Sys_{latency}$ \\
    }
}

\SetKwFunction{Ftwo}{Weight\_Locality\_Opt}
\SetKwProg{Fn}{Function}{:}{}
\Fn{\Ftwo{}}{
    \texttt{Knapsack\_Solver}($G_{model}$, $G_{sys}$) \\
    $Sys_{latency}, Sys_{energy} \longleftarrow$ \texttt{update\_System\_Scheduling()} 
}

\SetKwFunction{Fthree}{Activation\_Transfer\_Opt}
\SetKwProg{Fn}{Function}{:}{}
\Fn{\Fthree{}}{
    \For{every node pair adjacent in $G_{sys}$ }{
        \texttt{activation\_Fusion}(\textit{node pair}) \\
    }
    $Sys_{latency}, Sys_{energy} \longleftarrow$ \texttt{update\_System\_Scheduling()} 
}

\SetKwFunction{Ffour}{Data\_Locality\_Remapping}
\SetKwProg{Fn}{Function}{:}{}
\SetKwRepeat{repeat}{Repeat}{Until}
\Fn{\Ffour{}}{
    \repeat{no more beneficial remapping operations}{
        \For{ every $n\in G_{model}$}{
            Attempt to remap $n$ to its neighbors' Acc\\
            \texttt{Weight\_Locality\_Opt()}\\
            \texttt{Activation\_Transfer\_Opt()}\\
            $\Delta Sys_{latency} \leftarrow$ \texttt{update\_System\_Scheduling()} \\
            Accept remap if $\Delta Sys_{latency}< 0$
        }
    }
}

\caption{H2H Mapping and Scheduling}
\label{alg:init}
\end{algorithm}

\vspace{-4pt}
\subsection{Weight Locality Optimization}
\label{sec:weight-locality}

\texttt{Weight_locality_Opt} is performed after computation-prioritized mapping by utilizing the local DRAM of each accelerator.
With local DRAM, weight transfer from main memory can be greatly reduced, and it is a common practice to buffer part (or all) of the weights of the DNN layer(s) \cite{guo2019dl}. 
In this system, since multiple layers are mapped to the same accelerator, the layer weights must be selectively stored in the local DRAM, under a certain memory budget. 
Therefore, we propose to use the Knapsack algorithm 
to store , as much as possible, weights in the accelerators' local DRAM to reduce weight transfer.
After pinning weights locally, we first update each layer's latency, and then update the system's scheduling and overall latency.
Note that, since changing the latency and scheduling of one layer can affect all its successor layers, we update the layer scheduling \textit{recursively}. This is especially efficient for graph structures, since in each iteration, we only update a node's direct successor neighbors without traversing the entire graph.
An example is shown in Fig.~\ref{Fig:algsteps} \circled{2}, where the colored blocks represent the layers whose weights are stored in the local DRAM with reduced latency. The system latency is also reduced.

\vspace{-0.3cm}
\subsection{Activation Transfer Optimization}
After weight locality optimization, the
activation (IFM/OFM) transfer will be optimized by
\texttt{Activation_Transfer_Opt}
to further reduce the overall cost.
This is based on the assumption that, if two adjacent layers are mapped to the same accelerator, their intermediate IFM and OFM can be reused locally by taking advantage of the local DRAM and thus the activation transfer from/to the main memory can be avoided. We call it activation fusion, which can be performed recursively, similar to the weight locality optimization:
for each mapped layer, it checks its successor neighbors for activation fusion, updates its own and its neighbors' latency, if applicable, and recursively updates the system's overall scheduling.
An example of activation fusion is shown in Fig. \ref{Fig:algsteps} \circled{3}, where the starred blocks indicate the layers that are applicable for fusion. 


        

\vspace{-4pt}
\subsection{Data Locality Aware Remapping}
The weight and activation optimization are post-optimizations for communication given a mapping solution.
In this step, we execute  \texttt{{Data_Locality_Remapping}}, for communication-oriented remapping, i.e., initial mapping tuning aiming at largely reducing communication cost.
Specifically, for each layer, we define a \textit{remapping} operation that re-allocates a layer from its source accelerator to a new destination accelerator, on which its predecessors and/or successors are mapped.
This remapping reduces the activation transfer time by allowing activation fusion, but may increase the computation latency.
The weight transfer latency may be increased or decreased depending on the available local DRAM capacity of the destination accelerator.
Therefore, to determine the exact effect of a remapping operation, weight locality and activation transfer optimization, i.e., step 2 and 3, must be re-executed for every remapping attempt.
We adopt a greedy algorithm and perform remapping attempt for every layer; a remapping is accepted only if it reduces the system's overall latency, i.e., the benefit of communication reduction outweighs the computation cost increment. The algorithm terminates when no more layers can be remapped with reduced overall latency.

An example of locality-aware remapping is shown in Fig. \ref{Fig:algsteps} \circled{4}. In this example, layer 3.1 is remapped from Acc2 to Acc1 since its neighbor layer 3.2 resides on Acc1, so that the activation transfer between layer 3.1 and layer 3.2 can be reduced. Although in this example, the scheduling of layer 2.2 cannot move earlier because of the layer dependency, in most cases, the source accelerator can reduce its latency because of the memory budgeted for weights is released and it can execute earlier because it can take the vacated cycles of the remapped layer.

\subsection{Extension for Dynamic Modality Change}
The proposed H2H algorithm can be easily extended to handle dynamic modality change scenarios
which is common in multi-sensor systems.
For instance, a health monitoring system may choose to turn on and off motion sensors based on the surrounding environment and the person's activity \cite{chen2018power}, and such dynamic modality change can be as frequent as several times within one second.
This motivates an extended H2H mapping for dynamic modality changes, where a new mapping with increased or decreased modalities (i.e., layers) depends on the previous mapping result to maximally re-use the buffered weights.
The advantage is to avoid weight loading for frequent modality change.

Therefore, we modify the proposed H2H algorithm as follows for dynamic modality change.
Given the previous mapping and weight buffering, for a new set of modalities (layers), it prioritizes the layer mapping if the layer's weights are already buffered on a certain accelerator. Then, we repeat steps 1 to 4 with a modified Knapsack algorithm, where part of the weight allocation is determined.

\section{Evaluations}

\subsection{Evaluation Settings}
\noindent\textbf{Heterogeneous models.} 
Table \ref{table:modality} summarizes 6 heterogeneous DNN models used in evaluation, spanning different domains including Augmented Reality (AR), Face Recognition, Sentiment Analysis, Activity Recognition, and Emotion Recognition. 
Most models use Convolution Neural Networks (ResNet, VGG, VD-CNN, and their variants) as backbones, and there are typically 3 to 5 backbones placed together for MMMT with cross-backbone data dependencies.

\noindent\textbf{Heterogeneous accelerators.} We survey 12 state-of-the-art FPGA-based Convolution/FC/LSTM accelerators and summarize them in Table~\ref{table:accs}. We replicate their performance models based on the original papers; we honor the local DRAM capacity $M_{acc}$ based on the FPGA boards used, ranging from 512 MB to 8 GB \cite{guo2019dl}.

\noindent\textbf{System modeling.} We modify MAESTRO~\cite{maestro} to a system-level infrastructure to model the cloud-scale multi-FPGA system as shown in Fig.~\ref{Fig:H2H} (bottom)~\cite{fowers2018configurable}. 
The Ethernet speed $BW_{acc}$ between FPGAs and main memory ranges from 1 G to 10 G Ethernet (0.125 to 1.25 GB/s) \cite{awsspeed}.
The system latency and energy are modeled based on the values reported in the accelerator papers in Table~\ref{table:accs}.

\begin{table}[htbp]
\tabcolsep 3pt
\vspace{-4pt}
\renewcommand\arraystretch{1.0}
\caption{Heterogeneous (MMMT) models}
\footnotesize
\label{table:modality}
\begin{tabular}{c|c|c|c}
\toprule
\textbf{Domain}      & \textbf{Model} & \textbf{Backbones}          & \textbf{Para.}\\ \midrule
Augmented Reality    & VLocNet \cite{vlocnet}        & ResNet-50 variants          & 192M \\ \hline
Face Recognition     & CASUA-SURF \cite{zhang2019dataset}     & ResNet-18 variants          & 13.2M\\ \hline
Sentiment Analysis   & VFS \cite{thuseethan2020multimodal}           & VGG and VD-CNN variants     & 365M\\ \hline
Face Recognition     & FaceBag  \cite{shen2019facebagnet}      & ResNet variants             & 25M\\ \hline
Activity Recognition & CNN-LSTM \cite{li2017concurrent}      & ConvNet and LSTM variants   & 16M\\ \hline
Emotion Recognition  & MoCap \cite{tripathi2018multi}          & Convolution and LSTM unit   & 8M\\
\bottomrule
\end{tabular}
\vspace{-4pt}
\end{table}

\begin{table}[htbp]
\tabcolsep 2.5pt
\renewcommand\arraystretch{1.0}
\caption{State-of-the-art FPGA DNN accelerators}
\footnotesize
\label{table:accs}
\begin{tabular}{c|c|c|c}
\toprule
\textbf{Name} & \textbf{Accelerator Type} & \textbf{Optimization} & \textbf{FPGA} \\ \midrule
J.Z \cite{zhang2017improving}                 & Convolution               & On-chip memory        & GX1150        \\ \hline
C.Z \cite{zhang2015optimizing}                  & Convolution               & Channel parallel.     & VC707         \\ \hline
W.J \cite{jiang2019achieving}                  & Convolution               & Memory and Channel    & ZCU102        \\ \hline
J.Q \cite{qiu2016going}                  & Conv/FC/(LSTM)            & Computing Generality  & ZC706         \\ \hline
A.C \cite{chang2017compiling}                  & Convolution               & Loop Optimization     & XC7Z045       \\ \hline
Y.G \cite{guan2017fp}                  & Conv/FC/LSTM              & Computing Generality  & Stratix-V     \\ \hline
T. M \cite{ma2017optimizing}                 & Convolution               & Loop Optimization     & GX1150        \\ \hline
A.P \cite{podili2017fast}                  & Convolution               & Winograd              & Stratix-V     \\ \hline
X.W \cite{wei2017automated}                  & Convolution               & Systolic Array        & GT1150        \\ \hline
S.H \cite{han2017ese}                  & LSTM/FC                   & Deep Pipeline         & XCKU060       \\ \hline
X.Z \cite{zhang2020achieving}                  & LSTM                      & Gate Parallelism      & PYNQ-Z1/VC707 \\ \hline
B.L \cite{li2020ftrans}                  & LSTM                      & Deep Pipeline         & VCU118        \\ \bottomrule
\end{tabular}
\vspace{-4pt}
\end{table}

\begin{figure*}[htbp]
\vspace{-20pt}
 \setlength{\belowcaptionskip}{-0.2cm}
  \centering
  \includegraphics[width=1\textwidth]{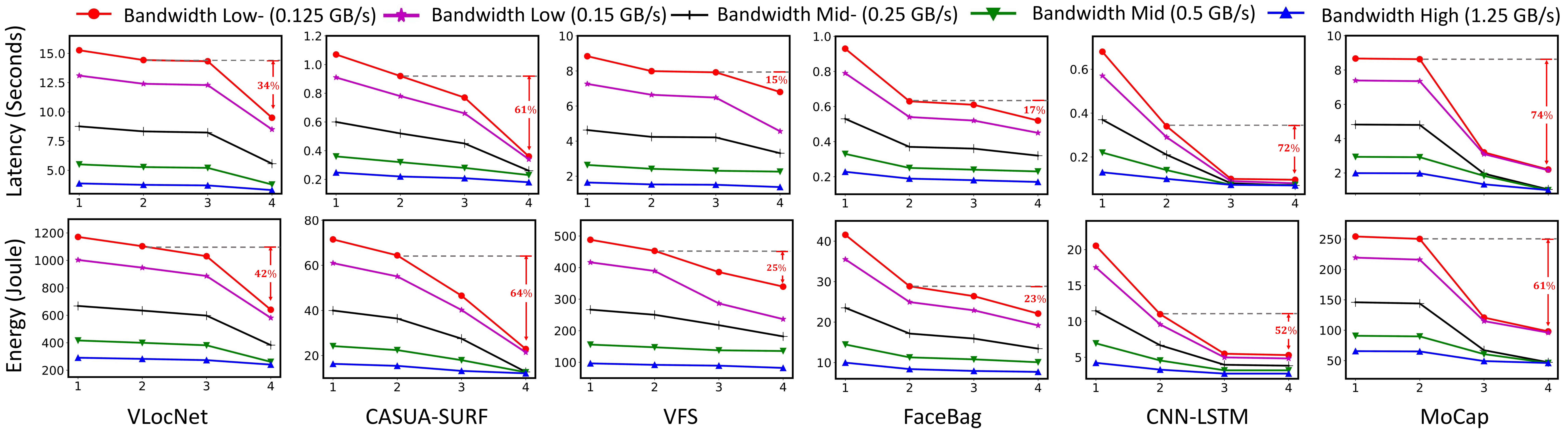}
  \caption{The latency and energy performance comparison.}\label{Fig:LatEnergy}
\end{figure*}


\subsection{H2H mapping performance}
\noindent \textbf{Baseline.}
As discussed in Section~\ref{sec:motivation},
existing mapping algorithms are computation-prioritized that strive for finding the most suitable accelerators based on the dataflow patterns~\cite{ kwon2021heterogeneous}.
This mapping strategy is the same as the first step in our H2H mapping.
To make a fair comparison, we take the results from H2H mapping after the second step (Section~\ref{sec:weight-locality}) including the weight locality optimization, since existing works can also assume local DRAM for the accelerators.

\noindent \textbf{Latency and Energy Reduction.} 
In Fig.~\ref{Fig:LatEnergy}, we present the latency and energy reduction of the 6 heterogeneous DNN models.
We test the H2H mapping algorithm under different network bandwidth configurations ($BW_{acc}$):
\texttt{Low-} (0.125 GB/s);
\texttt{Low} (0.15 GB/s);
\texttt{Mid-} (0.25 GB/s);
\texttt{Mid} (0.5 GB/s);
\texttt{High} (1.25 GB/s).
The x-axis refers to the four steps in H2H mapping algorithm, and the y-axis is the modeled system latency in seconds and energy in joule.
The H2H mapping algorithm achieves 15\% to 74\% system latency reduction and 23\% to 64\% energy reduction compared with the baseline mapping~\cite{kwon2021heterogeneous} when the system is bandwidth bounded, i.e., under the bandwidth \texttt{Low-} setting. 
With high bandwidth, the H2H still reduces overall latency by 10\% to 50\%.
In half of the evaluated cases, we achieved over 60\% latency reduction.

The detailed latency reduction after each step is shown in Table~\ref{table:latdetail}. 
Since we regard the second step as the baseline, we present the absolute latency values (in seconds) for steps 1 and 2 and the relative values for steps 3 and 4 compared with step 2.
Apparently, when the bandwidth increases, the reduction decreases, but even with high bandwidth, network CNN-LSTM and MoCap still reduce latency by almost half from H2H mapping.

\begin{figure}[htbp]
 \setlength{\belowcaptionskip}{-0.2cm}
  \centering
  \includegraphics[width=0.47\textwidth]{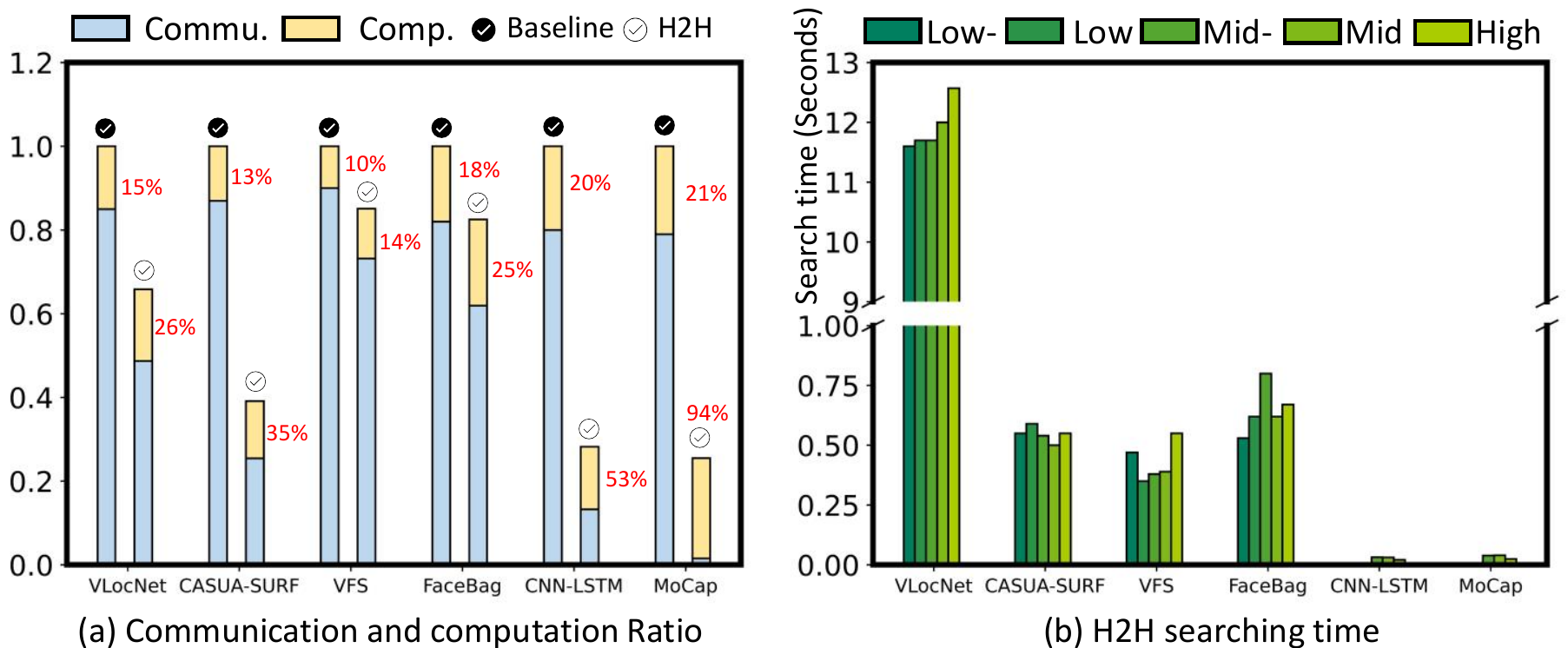}
  \caption{Communication and computation ratio. }\label{Fig:CTC}
\end{figure}

\noindent \textbf{H2H performance analysis.} In Fig.~\ref{Fig:CTC}(a), we visualize the communication and computation latency ratio using the mapping results under Bandwidth \texttt{Low-} of the six models. 
Note that after our H2H mapping, the computation ratio greatly increases (yellow bars), where MoCap increases from 21\% to 94\%, indicating that the communication overhead is largely reduced.
We also show the H2H mapping algorithm execution time in Fig.~\ref{Fig:CTC}(b). The search time is consistently low across different DNN models, less than one second. The VLocNet requires longer search time since it consists of 141 layers; the CNN-LSTM and MoCap are significantly faster since they consist of less than 30 layers.

\begin{table*}[htbp]
\tabcolsep 1.4pt
\renewcommand\arraystretch{1.0}
\caption{Latency reduction breakdown comparing with the second step (baseline).\vspace{-2pt}}
\footnotesize
\label{table:latdetail}
\begin{tabular}{c|cccc|cccc|cccc|cccc|cccc|cccc}
\toprule
 \multirow{ 2}{*}{\textbf{Bandwidth}} & \multicolumn{4}{c|}{\textbf{VLocNet}}                                                            & \multicolumn{4}{c|}{\textbf{CASUA-SURF}}                                                       & \multicolumn{4}{c|}{\textbf{VFS}}                                                              & \multicolumn{4}{c|}{\textbf{FaceBag}}                                                           & \multicolumn{4}{c|}{\textbf{CNN-LSTM}}                                                           & \multicolumn{4}{c}{\textbf{MoCap}}                                                            \\ \cmidrule{2-25}
  & \multicolumn{1}{c|}{\circled{1}}     & \multicolumn{1}{c|}{\circled{2}}     & \multicolumn{1}{c|}{\circled{3}}       & \multicolumn{1}{c|}{\circled{4}}       & \multicolumn{1}{c|}{\circled{1}}    & \multicolumn{1}{c|}{\circled{2}}    & \multicolumn{1}{c|}{\circled{3}}       & \multicolumn{1}{c|}{\circled{4}}       & \multicolumn{1}{c|}{\circled{1}}    & \multicolumn{1}{c|}{\circled{2}}    & \multicolumn{1}{c|}{\circled{3}}       & \multicolumn{1}{c|}{\circled{4}}       & \multicolumn{1}{c|}{\circled{1}}    & \multicolumn{1}{c|}{\circled{2}}     & \multicolumn{1}{c|}{\circled{3}}       & \multicolumn{1}{c|}{\circled{4}}       & \multicolumn{1}{c|}{\circled{1}}     & \multicolumn{1}{c|}{\circled{2}}     & \multicolumn{1}{c|}{\circled{3}}       & \multicolumn{1}{c|}{\circled{4}}       & \multicolumn{1}{c|}{\circled{1}}    & \multicolumn{1}{c|}{\circled{2}}    & \multicolumn{1}{c|}{\circled{3}}       & \multicolumn{1}{c}{\circled{4}}       \\ \midrule
\textbf{Low-} & \multicolumn{1}{c|}{15.27} & \multicolumn{1}{c|}{14.43} & \multicolumn{1}{c|}{99.31\%} & 65.84\% & \multicolumn{1}{c|}{1.07} & \multicolumn{1}{c|}{0.92} & \multicolumn{1}{c|}{83.70\%} & 39.13\% & \multicolumn{1}{c|}{8.84} & \multicolumn{1}{c|}{7.99} & \multicolumn{1}{c|}{99.12\%} & 85.11\% & \multicolumn{1}{c|}{0.93} & \multicolumn{1}{c|}{0.63}  & \multicolumn{1}{c|}{96.83\%} & 82.54\% & \multicolumn{1}{c|}{0.68} & \multicolumn{1}{c|}{0.34} & \multicolumn{1}{c|}{29.41\%} & 28.24\% & \multicolumn{1}{c|}{8.67} & \multicolumn{1}{c|}{8.63} & \multicolumn{1}{c|}{37.08\%} & 25.49\% \\ \hline
\textbf{Low}  & \multicolumn{1}{c|}{13.10}  & \multicolumn{1}{c|}{12.40}  & \multicolumn{1}{c|}{99.11\%} & 68.55\% & \multicolumn{1}{c|}{0.91} & \multicolumn{1}{c|}{0.78} & \multicolumn{1}{c|}{84.62\%} & 43.59\% & \multicolumn{1}{c|}{7.26} & \multicolumn{1}{c|}{6.64} & \multicolumn{1}{c|}{97.59\%} & 68.67\% & \multicolumn{1}{c|}{0.79} & \multicolumn{1}{c|}{0.54}  & \multicolumn{1}{c|}{96.30\%} & 83.33\% & \multicolumn{1}{c|}{0.57} & \multicolumn{1}{c|}{0.29}  & \multicolumn{1}{c|}{31.03\%} & 27.59\% & \multicolumn{1}{c|}{7.39} & \multicolumn{1}{c|}{7.35} & \multicolumn{1}{c|}{42.18\%} & 29.39\% \\ \hline
\textbf{Mid-}  & \multicolumn{1}{c|}{8.76}  & \multicolumn{1}{c|}{8.33}  & \multicolumn{1}{c|}{98.80\%} & 66.87\% & \multicolumn{1}{c|}{0.60}  & \multicolumn{1}{c|}{0.52} & \multicolumn{1}{c|}{86.54\%} & 50.00\% & \multicolumn{1}{c|}{4.63} & \multicolumn{1}{c|}{4.24} & \multicolumn{1}{c|}{99.29\%} & 78.07\% & \multicolumn{1}{c|}{0.53} & \multicolumn{1}{c|}{0.37}  & \multicolumn{1}{c|}{97.30\%} & 86.49\% & \multicolumn{1}{c|}{0.37}  & \multicolumn{1}{c|}{0.21}  & \multicolumn{1}{c|}{38.10\%} & 33.33\% & \multicolumn{1}{c|}{4.82} & \multicolumn{1}{c|}{4.8}  & \multicolumn{1}{c|}{40.83\%} & 21.67\% \\ \hline
\textbf{Mid}  & \multicolumn{1}{c|}{5.51}  & \multicolumn{1}{c|}{5.28}  & \multicolumn{1}{c|}{98.67\%} & 71.78\% & \multicolumn{1}{c|}{0.36} & \multicolumn{1}{c|}{0.32} & \multicolumn{1}{c|}{87.50\%} & 71.88\% & \multicolumn{1}{c|}{2.64} & \multicolumn{1}{c|}{2.42} & \multicolumn{1}{c|}{95.45\%} & 93.39\% & \multicolumn{1}{c|}{0.33} & \multicolumn{1}{c|}{0.25}  & \multicolumn{1}{c|}{96.00\%} & 92.00\% & \multicolumn{1}{c|}{0.22}  & \multicolumn{1}{c|}{0.14}  & \multicolumn{1}{c|}{52.14\%} & 52.14\% & \multicolumn{1}{c|}{2.94} & \multicolumn{1}{c|}{2.92} & \multicolumn{1}{c|}{62.67\%} & 34.25\% \\ \hline
\textbf{High} & \multicolumn{1}{c|}{3.88}  & \multicolumn{1}{c|}{3.76}  & \multicolumn{1}{c|}{98.67\%} & 88.03\% & \multicolumn{1}{c|}{0.25} & \multicolumn{1}{c|}{0.22} & \multicolumn{1}{c|}{94.55\%} & 81.82\% & \multicolumn{1}{c|}{1.64} & \multicolumn{1}{c|}{1.53} & \multicolumn{1}{c|}{98.69\%} & 90.20\% & \multicolumn{1}{c|}{0.29} & \multicolumn{1}{c|}{0.189} & \multicolumn{1}{c|}{95.24\%} & 89.95\% & \multicolumn{1}{c|}{0.13}  & \multicolumn{1}{c|}{0.10}  & \multicolumn{1}{c|}{73.00\%} & 70.00\% & \multicolumn{1}{c|}{1.99} & \multicolumn{1}{c|}{1.98} & \multicolumn{1}{c|}{67.68\%} & 50.51\% \\ \bottomrule
\end{tabular}
\end{table*}



\section{Conclusion}
In this work, we proposed a computation and communication aware H2H algorithm, 
aiming at system-level formulation, modeling, and optimization for mapping heterogeneous models to heterogeneous systems.
We achieve up to 74\% system latency reduction and 64\% energy reduction.
The H2H is designed with high flexibility, as it is configurable at system level and adopts a plug-in manner for accelerators. It can be easily configured to catch up with the latest advancements in deep learning society, such as the growing size of DNN models, increasing intensity of accelerator computing resource, and system network bandwidth. 





\bibliographystyle{unsrt}


\end{document}